\documentclass{article}

\usepackage{ijcai19}

\usepackage{graphicx}
\usepackage{array}
\usepackage{multirow}
\usepackage{multicol} 
\usepackage{mathrsfs}
\usepackage{algorithmic}
\usepackage[inoutnumbered,linesnumbered,ruled,vlined]{algorithm2e}
\usepackage{epsfig}
\usepackage{epstopdf}
\usepackage{subfigure}
\usepackage{enumerate}
\usepackage{times}
\usepackage{helvet}
\usepackage{courier}
\usepackage{url}
\usepackage{amsmath,amssymb,amsthm}
\usepackage{color}
\usepackage[small]{caption}
\usepackage{xfrac}
\usepackage{bbding}
\usepackage{pifont}
\usepackage{rotating}
\usepackage[table]{xcolor}
\usepackage{mathtools}
\usepackage{bm}

\newtheorem{defn}{Definition}

\newtheorem{ex}{Example}

\newcommand{\ginv}[2]{\mathcal{G}^{#1,#2}_{inv}}

\newcommand{\gvar}[2]{\mathcal{G}^{#1,#2}_{var}}

\def\Plus{\texttt{+}}

\newcommand{\skipNow}[1]{}
\newcommand{\todo}[1]{{\color{black}#1}}
\newcommand{\change}[1]{{\color{black}#1}}
\newcommand{\tocheck}[1]{{\color{black}#1}}
\newcommand{\jy}[1]{{\color{black}#1}}
\newcommand{\freddy}[1]{{\color{black}#1}}
\newcommand{\jeff}[1]{{\color{black}#1}}
\newcommand{\fl}[1]{{\color{black}#1}} 
\newcommand{\jp}[1]{{\color{black}#1}} 
\newcommand{\cjy}[1]{{\color{black}#1}}

\DeclareMathAlphabet{\mathpzc}{OT1}{pzc}{m}{it}
\DeclareMathAlphabet\mathbfcal{OMS}{cmsy}{b}{n}

\begin{document}

\title{Augmenting Transfer Learning with Semantic Reasoning
}



\author{
Freddy L\'ecu\'e$^{1,2}$ \and Jiaoyan Chen$^{3}$ \and Jeff Z. Pan$^{4,5}$ and Huajun Chen$^{6,7}$
\affiliations
$^1$CortAIx Thales, Montreal, Canada \\
$^2$Inria, Sophia Antipolis, France \\
$^3$Department of Computer Science, University of Oxford, UK\\
$^4$Department of Computer Science, The University of Aberdeen, UK\\
$^5$Edinburgh Research Centre, Huawei, UK\\
$^6$College of Computer Science, Zhejiang University, China \\
$^7$ZJU-Alibaba Joint Lab on Knowledge Engine, China
}

\maketitle

\begin{abstract}
Transfer learning aims at building \change{robust prediction models} by transferring knowledge gained from
\cjy{one problem to another.} 
\change{
In the semantic Web, learning tasks are enhanced with semantic representations.
\cjy{We exploit their semantics to augment transfer learning by 
dealing with \textit{when to transfer} with semantic measurements and \textit{what to transfer} with semantic embeddings.
We further present a general framework that integrates the above measurements and embeddings with existing transfer learning algorithms for higher performance.
It has demonstrated to be robust in two real-world applications: bus delay forecasting and air quality forecasting. }
}
\end{abstract}

\section{Introduction}

Transfer learning \cite{pan2010survey} 
aims at solving the problem of lacking training data by utilizing data from other related \jy{\textit{learning domains}, each of which is referred to as a pair of dataset and prediction task.}
\freddy{Transfer Learning plays a critical role in real-world applications of ML as (labelled) data is usualy not large enough to train accurate and robust models.}
Most approaches focus on similarity in raw data distribution with techniques such as dynamic weighting of instances \cite{dai2007boosting} and 
model parameters sharing \cite{benavides2017accgensvm} (cf. Related Work).

Despite of a large spectrum of techniques \cite{weiss2016survey} in transfer learning, \freddy{it remains challenging to assess a priori which domain and data set to elaborate from \cite{dai2009translated}. 
To deal with such challenges, }
\cite{choi2016knowledge} integrated expert feedback as semantic representation on 
\jy{domain}
similarity for knowledge transfer
while \cite{lee2017transfer} evaluated the graph-based representations of source and target domains. 
Both studies encode semantics but are limited by the expressivity,
\freddy{which 
restricts domains interpretability and inhibits a good understanding of transferability.
\jp{There are also efforts on  Markov Logic Networks (MLN) based transfer learning, by using  first order~\cite{MHM2007,MiMo2009} or second order~\cite{DaDo2009,HKD2015} rules as declarative   prediction models. However, these efforts still cannot answer questions like:
What ensures a positive domain transfer? 
Would learning a model from road traffic congestion in London be the best for predicting congestion in Paris? Or would an air quality model transfer better?} 
}

\jp{In this paper, we propose to encode the semantics of learning tasks and domains with 
OWL ontologies
and provide a robust foundation to study transferability between source and target learning domains}. 
%
%
From knowledge materialization \cite{nickel2016review}, feature selection \cite{vicient2013automatic}, predictive reasoning \cite{lecue2015consistent}, stream learning \cite{chen2017learning} \jy{to transfer learning explanation \cite{DBLP:conf/kr/ChenLPHC18}}, 
all are examples of inference tasks where the semantics of data representation are exploited for deriving a priori knowledge from pre-established 
statements in \jy{ML} tasks.

\begin{figure}[h]
\centering
\includegraphics[scale=0.3]{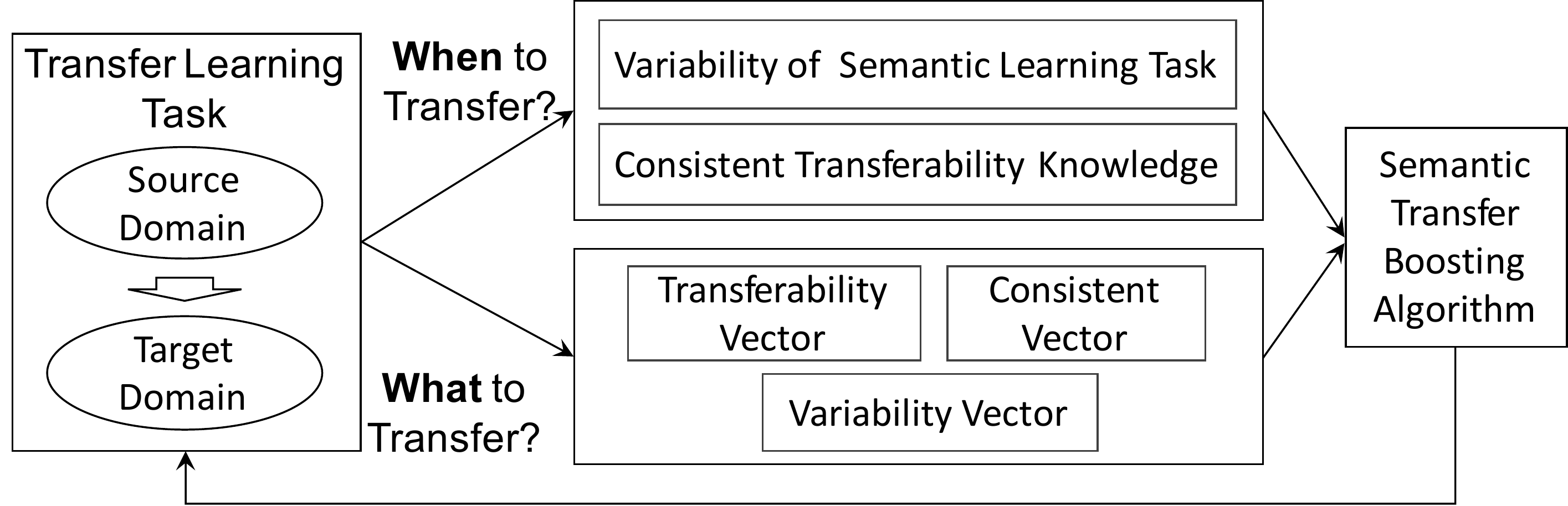}
\caption{Ontology-based Transfer Learning Augmentation.}
\label{fig:framework}
\end{figure}

\cjy{We introduce a framework to augment transfer learning by semantics and its reasoning capability, as shown in Figure \ref{fig:framework}.
It deals with \textit{(i)} \emph{when to transfer} by suitable transferability measurements (i.e., variability of semantic learning task and consistent transferability knowledge),
\textit{(ii)} \emph{what to transfer} by embedding the semantics of learning domains and tasks with transferability vector, consistent vector and variability vector.
In addition to expose semantics that drives transfer, 
a transfer boosting algorithm is developed to integrate the embeddings with existing transfer learning approaches.
}
Our approach achieves high performance for 
multiple transfer learning tasks in air quality and bus delay forecasting.
%

\section{Background}{\label{sec:Background}}

%
%
%
Our work uses OWL ontologies underpinned by
Description Logic (DL) $\mathcal{EL}^{++}$ \cite{BaaBL05}\cite{bechhofer2004owl} \cjy{to model the semantics of learning domains and tasks}.
%
%

\subsection{Description Logics $\mathcal{EL}^{++}$ and Ontology}

A signature $\Sigma$, noted 
$(\mathcal{N}_C, \mathcal{N}_R, \mathcal{N}_I)$ consists of $3$ disjoint sets of \textit{(i)} atomic concepts $\mathcal{N}_C$, \textit{(ii)} atomic roles $\mathcal{N}_R$, and \textit{(iii)} individuals $\mathcal{N}_I$. 
Given a signature, the top concept $\top$, the bottom concept $\bot$, an atomic concept $A$, an individual $a$, an atomic role expression $r$, $\mathcal{EL}^{++}$ concept expressions $C$ and $D$ in $\mathcal{C}$ can be composed with the following constructs:
\begin{equation}
\top\;|\;\bot\;|\;A\;|\;C\sqcap D\;|\;\exists r.C\;|\;\{a\}\nonumber
\end{equation}
%
%
A DL
ontology is composed of 
a TBox $\mathcal{T}$ and 
an ABox $\mathcal{A}$. 
$\mathcal{T}$ is a set of concept,
role axioms. $\mathcal{EL}^{++}$ supports General Concept Inclusion axioms (GCIs e.g., $C \sqsubseteq D$), 
Role Inclusion axioms (RIs e.g., $r \sqsubseteq s$
).
$\mathcal{A}$ is a set of class assertion axioms, e.g., $C(a)$, role assertion axioms, e.g., $r(a, b)$, 
individual in/equality axioms e.g., $a \neq b$,
$a = b$. \jeff{
Given an input ontology $\mathcal{T} \cup \mathcal{A}$, we consider  the closure of atomic  ABox entailments  (or simply entialment closure, denoted as $\mathcal{G}(\mathcal{T \cup A})$) as $\left\{g | \mathcal{T} \cup \mathcal{A} \models g \right\}$,
where $g$ represents an  atomic concept assertion $A(b)$, or an atomic role assertion  entailment $r(a,b)$, involving only named concepts, named roles and named individuals.}
%
%
\jy{Entailment reasoning in $\mathcal{EL}^{++}$ is PTime-Complete.}

\begin{figure}[h]
\centering
\includegraphics[scale=0.185]{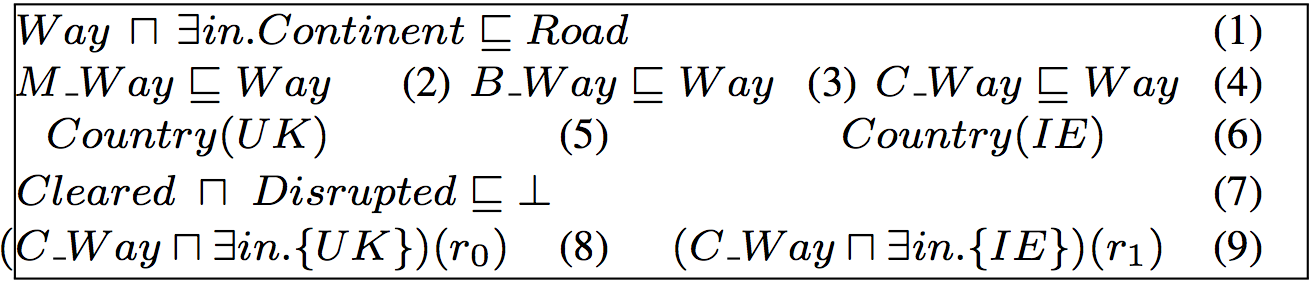}
\caption{Sample of an Ontology's TBox $\mathcal{T}$ and ABox $\mathcal{A}$.}
\label{fig:StaticOntology:Background:Knowledge}
\end{figure}

\begin{ex}\label{ex:ex1}\textbf{(TBox and ABox Concept Assertion Axioms)}\\
\change{
Figure \ref{fig:StaticOntology:Background:Knowledge} presents (i) a TBox $\mathcal{T}$ where $Road$ (1) 
denotes the concept of ``{ways which are in a continent}", and (ii) concept assertions 
(8-9)
with individuals $r_0$ and $r_1$ being roads.}
\end{ex}
%

\subsection{Learning Domain and Task}
\cjy{To model the learning domain with ontology, we use Learning Sample Ontology and Target Entailment, as in \cite{DBLP:conf/kr/ChenLPHC18}.
A learning domain \jp{consists of} an LSO set (i.e., dataset) and a target entailment set (i.e., prediction task).}

\begin{defn}{\label{defn:as}}\textbf{(Learning Sample Ontology (LSO))}\\
A learning sample ontology  
$\mathcal{O} = \langle \langle \mathcal{T}, \mathcal{A} \rangle, S \rangle$
is an ontology $\langle \mathcal{T}, \mathcal{A} \rangle$ annotated by property-value pairs $S$. 
\end{defn}

The annotation $S$ 
acts as key dimensions to uniquely identify an input sample of ML methods. When the context is clear, we also use LSO to refer to its ontology  $\langle \mathcal{T}, \mathcal{A} \rangle$.

\todo{
\begin{ex}{\label{ex:lso}}\textbf{(An LSO in Context of Ireland Traffic)}\\
Assume an LSO is annotated by property-value pairs $S:=\{$topic: \freddy{Road}, \freddy{$road:$ C\_Way}, \freddy{$country:$ UK$\}$}.
Its TBox $\mathcal{T}$ includes static axioms like 
(1);
its ABox $\mathcal{A}$ includes facts e.g., $hasAvgSpeed(r_0, Low)$ that are observed in $C\_Way$ in UK.
\end{ex}
}


\begin{defn}
{\label{defn:mld}}\textbf{(Learning Domain and Target Entailment)}\\
A learning domain $\mathcal{D} = \langle  \mathbb{O}, \mathcal{G}^{\mathcal{Y}} \rangle$ consists of 
a set of LSOs $\mathbb{O}$  that share the same TBox $\mathcal{T}$,
and target entailments $\mathcal{G}^{\mathcal{Y}}$, each of whose 
truth in an LSO
is to be predicted.
Its entailment closure, denoted as $\mathcal{G}(\mathbb{O})$ is defined as $\cup_{\mathcal{O} \in \mathbb{O}} \mathcal{G}(\mathcal{O})$.
\end{defn}

Definition \ref{defn:lwo} revisits supervised learning within a domain.
In a training LSO,
a target entailment is true if it is entailed by an LSO, and false otherwise.
\cjy{In a testing LSO,
the truth of a target entailment is to be predicted instead of being inferred.}

%
\begin{defn}{\label{defn:lwo}}\textbf{(\freddy{Semantic Learning Task)}}\\
Given a learning domain $\mathcal{D} = \langle \mathbb{O}, \mathcal{G}^{\mathcal{Y}} \rangle$, 
whose LSOs $\mathbb{O}$ are divided into two disjoint sets $\mathbb{O}^{\prime}$ and $\mathbb{O}^{\prime\prime}$,
a \freddy{semantic learning task}, denoted by $\mathbb{T}$, within $\mathcal{D}$, \freddy{is defined as:} $\langle \mathcal{D}, \mathbb{O}^{\prime}, \mathbb{O}^{\prime\prime}, f(\cdot) \rangle$ \freddy{i.e.,  the} task of
identifying a function $f(\cdot)$ with $\mathbb{O}^{\prime}$ and $\mathcal{G}^{\mathcal{Y}}$ 
to predict 
the truth of $\mathcal{G}^{\mathcal{Y}}$ in each $\mathcal{O}$ in $\mathbb{O}^{\prime\prime}$.
Here, $\mathbb{O}^{\prime}$ is called a training LSO set, 
while $\mathbb{O}^{\prime\prime}$
is called a testing LSO set.
\end{defn}

\begin{ex}{\label{ex:mld}}\textbf{(Semantic Learning Task)}\\
Given a domain composed of LSOs annotated by $\{topic:$ Road, $country:$ UK$\}$ and target entailments $Cleared(r_0)$ and $Disrupted(r_0)$,
 the LSOs are divided into a training set $\mathbb{O}^{\prime}$ and a testing set $\mathbb{O}^{\prime\prime}$ according to \freddy{the type of roads involved}, \freddy{the objective} is to identify a function from $\mathbb{O}^{\prime}$ to predict
 the condition of road $r_0$, namely 
 the truth of $Cleared(r_0)$ and $Disrupted(r_0)$ in each LSO in $\mathbb{O}^{\prime\prime}$.
\end{ex}



\subsection{Transfer Learning Across Domains}
\cjy{Definition \ref{defn:ont_tr} revisits \textit{transfer learning}
where $\mathcal{D}_{S}$ and $\mathcal{D}_{T}$ are called \textit{source domain} and \textit{target domain} and 
their entailment closures 
are denoted as $\mathcal{G}_S$ and $\mathcal{G}_T$.}

\begin{defn}{\label{defn:ont_tr}}\textbf{(Transfer Learning)}\\
Given two learning domains 
$\mathcal{D}_{S}$ $=$ $\langle  \mathbb{O}_{S}, \mathcal{G}_{S}^{\mathcal{Y}}\rangle$ and $\mathcal{D}_{T}$ $=$ $\langle \mathbb{O}_{T}, \mathcal{G}_{T}^{\mathcal{Y}} \rangle$, 
where the LSOs of $\mathcal{D}_{T}$ are divided into two disjoint sets $\mathbb{O}_{T}^{\prime}$ and $\mathbb{O}_{T}^{\prime\prime}$,
transfer learning from $\mathcal{D}_{S}$ to $\mathcal{D}_{T}$ is a 
task of learning a prediction function $f_{T|_{\mathcal{S}}}(\cdot)$ from 
$\mathbb{O}_{S}$, $\mathcal{G}_{S}^{\mathcal{Y}}$, $\mathbb{O}_{T}^{\prime}$ and $\mathcal{G}_{T}^{\mathcal{Y}}$
to predict 
the truth of $\mathcal{G}_{T}^{\mathcal{Y}}$ in each LSO in $\mathbb{O}_{\beta}^{\prime\prime}$.
\end{defn}

\todo{
\begin{ex}{\label{ex:TransferLearning}}\textbf{(Transfer Learning)}\\ 
\cjy{Assume $\mathcal{D}_T$ is the domain in Example \ref{ex:mld},
$\mathcal{D}_S$ is a domain with LSOs annotated by $\{topic$: Road, $country$: IE$\}$,
an example of transfer learning is to identify a function  using all the LSOs of Dublin traffic and the training LSOs of London traffic ($\mathbb{O}_T^{\prime}$) for predicting the traffic condition of road $r_0$ in each testing LSO of London traffic ($\mathbb{O}_T^{\prime\prime}$).}
%
\end{ex}
}


\cjy{We demonstrate how ontology-based descriptions 
can drive transfer learning from one domain to another.
To this end, similarities between domains are first  characterized.
We adopt the
variability of ABox entailments \cite{DBLP:conf/ijcai/Lecue15} in Definition \ref{defn:streamSimilarity},
where \eqref{eq:variantAxioms} reflects $var$iant knowledge between two domains while  
\eqref{eq:invariantAxioms} denotes $inv$ariant knowledge.}

\begin{figure}[h]
\centering
\includegraphics[scale=0.185]{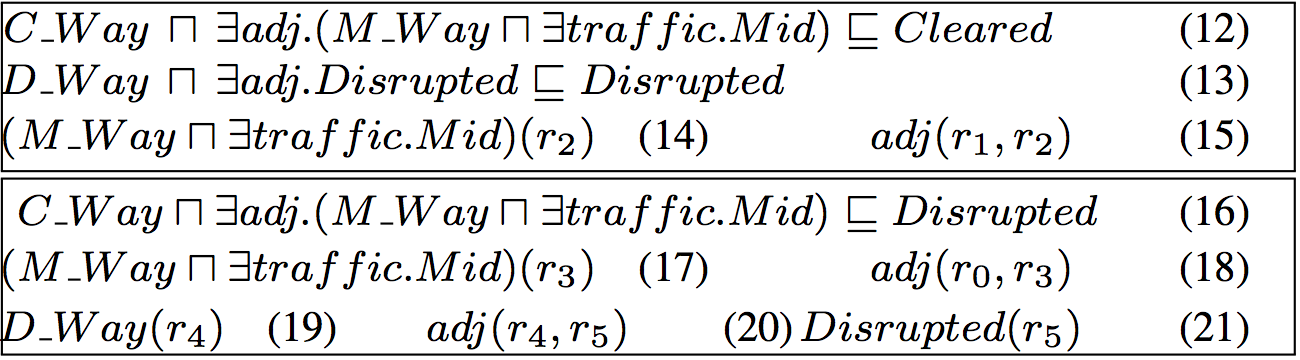}
\caption{[Up] Source Domain Ontologies $\mathbb{O}_{S}$ in Context of IE Traffic; [Down] Target Domain Ontologies $\mathbb{O}_{T}$ in Context of UK Traffic.}
\label{fig:OAlphaBeta}
\end{figure}

%
\begin{defn}{\label{defn:streamSimilarity}}\textbf{(Entailment-based 
Domain
\change{Variability})}\\
Given a source learning domain $\mathcal{D}_S$ and a target learning domain $\mathcal{D}_T$,
let $\mathcal{G} = \mathcal{G}_{S} \cup \mathcal{G}_{T}$,
%
the variability from $\mathcal{D}_{S}$ to
$\mathcal{D}_{T}$, denoted as $\nabla(\mathcal{O}_{T}, \mathcal{O}_{S})$
are ABox entailments:
\setcounter{equation}{9}
\begin{small}
\begin{align}
\gvar {[S]} {[T]} & = \{g\in\mathcal{G}\;|\; g \in \mathcal{G}_{T} \;\vee g \not \in \mathcal{G}_{S} \}\label{eq:variantAxioms}\\
\ginv {[S]} {[T]} & = \{g\in\mathcal{G}\;|\; g \in \mathcal{G}_{T} \;\wedge g \in \mathcal{G}_{S} \}\label{eq:invariantAxioms}
\end{align}
\end{small}
\end{defn} 
\begin{ex}{\label{ex:streamSimilarity}}\textbf{(Entailment-based 
Domain \change{Variability})}\\
Let Figure \ref{fig:OAlphaBeta}, which capture the contexts in IE and UK\fl{,} be ontologies of $\mathcal{D}_S$ and $\mathcal{D}_T$ respectively.
Table \ref{tab:streamSimilarity} illustrates some variabilities of $\mathcal{D}_{S}$ 
and $\mathcal{D}_{T}$ through ABox entailments. 
For instance $r_1$ as a disrupted road in $\mathcal{D}_{S}$ is $new$ ($variant$) w.r.t. knowledge in $\mathcal{D}_{T}$ and axioms 
(1), (9) 
and (12-15). 
\end{ex}

\begin{table}[h!]
\begin{smallermathTable}
\centering
\begin{tabular}[t]{@{ }l@{ }|p{1.8cm}<{\centering}|p{1.8cm}<{\centering}}
\hline
 \multirow{2}{*}{Ontology Variability} & \multicolumn{2}{|c}{$\nabla(\mathcal{D}_{S},\mathcal{D}_{T})$} \\
& $variant$ & $invariant $ \\\hline
$Road(r_3)$ & & \checkmark \\
$Cleared(r_1)$  & \checkmark &\\
$Disrupted(r_0)$  & \checkmark &\\

\end{tabular}
\caption{\label{tab:streamSimilarity}
Examples for Entailment-based 
Domain
Variability.
}
\end{smallermathTable}
\end{table}
\section{\change{Transferability}}\label{sec:AAT}

We present \textit{(i)} variability of semantic learning tasks, 
\textit{(ii)} semantic transferability, as a basis for qualifying,
quantifying transfer learning 
(i.e., \emph{when to transfer}), together with \textit{(iii)} indicators (i.e., \emph{what to transfer}) driving transferability.
\fl{They 
are pivotal properties, as any change in domains, their transfer function and consistency drastically impact the quality of derived models \cite{DBLP:conf/icml/LongC0J15,DBLP:conf/kr/ChenLPHC18}.}

\subsection{Variability of Semantic Learning Tasks}

\todo{Definition \ref{defn:domainVariability} extends \emph{entailment-based ontology variability} (Definition \ref{defn:streamSimilarity}) 
to capture the learning task variability,
where 
$(\cdot) ^ {[\mathcal{Y}_S],[\mathcal{Y}_T]}$ represents using target entailments in \eqref{eq:variantAxioms} \eqref{eq:invariantAxioms}.
}

\begin{defn}{\label{defn:domainVariability}}\textbf{(Variability of Semantic Learning Tasks)}\\
Let $\mathbb{T}_S$ and $\mathbb{T}_T$ be semantic learning tasks of source learning domain $\mathcal{D}_S$ and target learning domain $\mathcal{D}_T$.
%
The variability of semantic learning tasks
$\nabla(\mathbb{T}_{S}, \mathbb{T}_T)$ 
is defined by \eqref{eq:domainVariability}, where
$|\cdot|$ refers to the cardinality of a set.
\setcounter{equation}{21}
\begin{small}
\begin{equation}
\left(
\frac{|\gvar {[S]} {[T]}|}{|\gvar {[S]} {[T]}| + |\ginv {[S]} {[T]}|},
\frac{|\gvar {[\mathcal{Y}_S]} {[\mathcal{Y}_T]}|}{|\gvar {[\mathcal{Y}_S]} {[\mathcal{Y}_T]}| + |\ginv {[\mathcal{Y}_S]} {[\mathcal{Y}_T]}|}\right)
\label{eq:domainVariability}
\end{equation}
\end{small}
%
%
%
%
\end{defn}
%
%

The variability of semantic learning tasks 
\eqref{eq:domainVariability}, also represented by $(\nabla(\mathbb{T}_{S},\mathbb{T}_{T})|_{\mathbb{O}},\nabla(\mathbb{T}_{S},\mathbb{T}_{T})|_{\mathcal{Y}})$
in $[0,1]$, captures the variability of 
source and \tocheck{target} domain LSOs as well as the variability of target entailments.
The higher values the stronger variability. 
%
The calculation of \eqref{eq:domainVariability} is in worst case polynomial time w.r.t size $\mathbb{O}_S$, $\mathbb{O}_T$, $\mathcal{Y}_S$,
$\mathcal{Y}_T$ in $\mathcal{EL}^{++}$. 
Its evaluation requires (i) ABox entailment, 
(ii) basic set theory operations from Definition \ref{defn:streamSimilarity}, both in polynomial time 
\cite{BaaBL05}.

\begin{ex}{\label{ex:domainVariability}}\textbf{(Variability of Semantic Learning Tasks)}\\
The variability of learning task between $\mathbb{T}_S$ and $\mathbb{T}_T$ in Example \ref{ex:TransferLearning} is $(\sfrac{2}{3}, 0)$ as the number of variant and invariant ABox entailments are respectively $6$ and $3$, and $\mathcal{Y}_S = \mathcal{Y}_T$.
\fl{i.e., moderate variability of domains, none for target variables.}
\end{ex}

\subsection{Semantic Transferability - When to Transfer?}

We define \emph{semantic transferability} 
from a source to a target semantic learning task as the existence of knowledge
that are captured as ABox entailments~\cite{PaTh07} in the source
and have positive effects on predictive quality of the prediction function of the target semantic learning task. 

%
\begin{defn}{\label{defn:semanticTransferability}}\textbf{(Semantic $\varepsilon$-Transferability)}\\
Let $\mathbb{T}_S$,
$\mathbb{T}_T$ be source, target semantic learning 
tasks with 
entailment closures $\mathcal{G}_S$, $\mathcal{G}_T$.
Semantic $\varepsilon$-transferability $\mathbb{T}_S \overset{\varepsilon} \mapsto \mathbb{T}_T$ occurs from $\mathbb{T}_S$ to $\mathbb{T}_T$ 
iff $\exists \mathcal{S} \subseteq \mathbb{O}_S:$
\begin{small}
\begin{multicols}{2}\noindent
\begin{equation}
{{m}}(f_{T|
_{\mathcal{S}}}(\cdot)) - {{m}}(f_T(\cdot)) > \varepsilon \label{eq:betterquality}\\ 
\end{equation}
\begin{equation}
\mathcal{G}_{\mathcal{S}} \neq \mathcal{G}_T \label{eq:newKnowledge}
\end{equation}
\end{multicols} 
\end{small}
\noindent where $f_{T|_{\mathcal{S}}}(\cdot)$ is the predictive function $f_T(\cdot)$ w.r.t. $\mathbb{O}_T\cup\mathcal{S}$. $\mathcal{G}_{\mathcal{S}}$ is the ABox closures of $\mathcal{S}$.
%
%
%
\end{defn}
\noindent ${\mathcal{S}}$ is 
knowledge from $\mathbb{O}_S$, 
to be used for over-performing the predictive quality of $f_T(\cdot)$ with a $\varepsilon\in(0,1]$ factor \eqref{eq:betterquality} while being new with respect to ABox entailments in $\mathcal{G}_{T}$ \eqref{eq:newKnowledge}.

\begin{ex}{\label{ex:semanticTransferabilityX}}\textbf{(Semantic $\varepsilon$-Transferability)}\\
Let $\mathbb{T}_S$, $\mathbb{T}_T$ \todo{be 
semantic learning tasks in $\mathcal{D}_S$, $\mathcal{D}_T$ in Example \ref{ex:TransferLearning},}
$\mathcal{S}$ be 
ABox entailment closure of 
(12-15)
in $\mathcal{O}_S$, and 
${{m}}(f_{T|_{\mathcal{S}}}(\cdot)) > {{m}}(f_T(\cdot))$. 
%
Semantic $\varepsilon$-transferability occurs from $\mathbb{T}_S$ to $\mathbb{T}_T$ as (i) an $\varepsilon > 0$, satisfying condition \eqref{eq:betterquality}, exists, and (ii) 
\eqref{eq:newKnowledge} is true 
cf. Table \ref{tab:streamSimilarity} w.r.t. $\mathcal{S}$.
%
\todo{Thus, knowledge $\mathcal{S}$ in IE traffic context ($\mathcal{D}_S$) ensures transferability from $\mathcal{D}_S$ to $\mathcal{D}_T$ for traffic prediction in UK.}
\end{ex}

ABox entailments $\mathcal{S}$ satisfying Definition \ref{defn:semanticTransferability} are denoted as \emph{transferable knowledge} while those contradicting \eqref{eq:betterquality} i.e., ${{m}}(f_{T|_{\mathcal{S}}}(\cdot)) - {{m}}(f_T(\cdot)) \leq \varepsilon$ are \emph{non-transferable knowledge} as they deteriorate predictive quality of 
target function $f_T(\cdot)$.

\begin{ex}{\label{ex:TransferableKnowledge}}\textbf{(Transferable Knowledge)}\\
Consider entailments in $\mathcal{S}$: (i) $Disrupted(r_4)$, derived from 
(13) (19-21),
(ii) $Cleared(r_0)$, derived from 
(8) (12) (17-18).
%
As part of knowledge $\mathcal{S}$ positively impacting the quality of the prediction task, they are also separate $\varepsilon$-transferable knowledge with max
$\varepsilon$:
$.1$,
$.07$ (computation details omitted).
\end{ex}

\subsection{Consistent Transferable Knowledge}

%
Transferring knowledge across domains can derive to inconsistency. Definition \ref{defn:consistentTransferableKnowledge} captures knowledge ensuring transferability 
while maintaining consistency in the target domain.

\begin{defn}{\label{defn:consistentTransferableKnowledge}}\textbf{(Consistent Transferable Knowledge)}\\
Let $\mathcal{S}$ be ABox entailments ensuring 
$\mathbb{T}_S \overset{\varepsilon} \mapsto \mathbb{T}_T$. 
$\mathcal{S}$ is consistent transferable knowledge from $\mathbb{T}_S$ to $\mathbb{T}_T$ iff $\mathcal{S} \cup \mathbb{O}_T \not\models \bot$.
\end{defn}

ABox entailments $\mathcal{S}$ satisfying $\mathcal{S} \cup \mathbb{O}_T \models \bot$ are called inconsistent transferable knowledge. 
They are interesting ABox entailments as they expose knowledge contradicting the target domain while maintaining transferability.
%
Evaluating if 
$\mathcal{S}$ is consistent transferable knowledge is in worst case polynomial time in $\mathcal{EL}^{++}$ w.r.t. 
size of 
$\mathcal{S}$ and 
$\mathbb{O}_T$. 

\begin{ex}{\label{ex:consistentTransferableKnowledge}}\textbf{((In-)Consistent Transferable Knowledge)}\\
$Disrupted(r_4)$ in $\mathcal{S}$ of Example \ref{ex:TransferableKnowledge} is consistent transferable knowledge in $\mathbb{T}_T$ as $\{Disrupted(r_4)\} \cup \mathbb{O}_T \not\models\bot$.
On contrary $Cleared(r_0)$ and 
$Disrupted(r_0)$ in $\mathcal{S}$, derived from (16-18)
are inconsistent 
(7).
Thus, $Cleared(r_0)$ in $\mathbb{O}_S$ is inconsistent transferable knowledge in $\mathbb{T}_T$.
\end{ex}

%


%
\section{Semantic Transfer Learning}

We tackle the problem of transfer learning by computing semantic embeddings (\emph{i.e., how to transfer}) for knowledge transfer, and determining a strategy to exploit the semantics of the learning tasks (Section \ref{sec:AAT}) in Algorithm \ref{algo:StAdaBoost}.

\subsection{Semantic Embeddings - How to transfer?}

The semantics of learning tasks exposes three levels of knowledge which are crucial for transfer learning:
%
variability,
%
transferability, 
%
consistency. 
They are encoded as embeddings through Definition \ref{def:transferabilityVector}, \ref{def:consistencyVector} and \ref{def:Weight}.

%
\begin{defn}{\label{def:transferabilityVector}}\textbf{(Transferability Vector)}\\
Let $\mathcal{G}=\{g_1,\ldots,g_m\}$ be all distinct ABox entailments in $\mathbb{O}_S\cup\mathbb{O}_T$.
A ${\bf t}$ransferability vector from $\mathbb{T}_S$ to $\mathbb{T}_T$, denoted by ${\bf t}(\mathcal{G})$, is a vector of dimension $m$ such that 
$\forall j\in[1,m]$:
%
%
\cjy{$t_{j} \stackrel{.}{=} \varepsilon_j$ if $g_j$ is $\varepsilon_j$-transferable knowledge, and $t_{j} \stackrel{.}{=} 0$ otherwise,}
\noindent with $\varepsilon_j\;|\;\nexists\varepsilon_j^{*}, \varepsilon_j < \varepsilon_j^{*}$ and $g$ is $\varepsilon_j^{*}$-transferable knowledge. 
\end{defn}

A transferability vector (Definition \ref{def:transferabilityVector}) is adapting the concept of feature vector \cite{bishop2006pattern} in Machine Learning to represent the qualitative transferability from source to target 
of all ABox entailments.
Each dimension captures the best of transferability of a particular ABox entailment.

\begin{ex}{\label{ex:transferabilityVector}}\textbf{(Transferability Vector)}\\
Suppose 
$\mathcal{G} \stackrel{.}{=} \{Disrupted(r_4), Cleared(r_0)\}$. Transferability vector ${\bf t}(\mathcal{G})$ is $(.1, .07)$ cf. $\varepsilon$-transferability in Example \ref{ex:TransferableKnowledge}.
\end{ex}

%
A consistency vector (Definition \ref{def:consistencyVector}) is computed from all entailments by 
evaluating their (in-)consistency, either $1$ or $0$, when transferred in the target semantic learning task.
%
%
Feature vectors are bounded to 
only raw data while transferability and consistency vectors, with 
larger dimensions, embed transferability and consistency of 
data and its inferred assertions. 
They ensure a larger,
more contextual coverage.
%
\begin{defn}{\label{def:consistencyVector}}\textbf{(Consistency Vector)}\\
Let $\mathcal{G}=\{g_1,\ldots,g_m\}$ be all distinct ABox entailments in $\mathbb{O}_S\cup\mathbb{O}_T$.
A ${\bf c}$onsistency vector from $\mathbb{T}_S$ to $\mathbb{T}_T$, denoted by ${\bf c}(\mathcal{G})$, is a vector of dimension $m$ such that 
$\forall j\in[1,m]$:
\cjy{
$c_{j} = 1$ if $\{g_j\}\cup\mathbb{O}_T\not\models\bot$, and $c_{j} = 0$ otherwise
}
\end{defn}

%

%
The variability vector (Definition \ref{def:Weight}) is used as an indicator of semantic variability between the two learning tasks. It is a value in $[0,1]$ with an emphasis on the domain ontologies and / or label space depending on its parameterization $(\alpha, \beta)$.
We characterize any variability weight above $\sfrac{1}{2}$ as \emph{inter-domain} transfer learning tasks, 
below $\sfrac{1}{2}$ as \emph{intra-domain}.

%
\begin{defn}{\label{def:Weight}}\textbf{(Variability Vector)}\\
Let $\mathcal{G}=\{g_1,\ldots,g_m\}$ be 
ABox entailments in $\mathbb{O}_S\cup\mathbb{O}_T$. 
%
A variability vector ${\bf v}(\mathcal{G},\alpha,\beta)$ from $\mathbb{T}_S$ to $\mathbb{T}_T$ is a vector of dimension $m$ 
with $\alpha, \beta \in [0,1]$ such that $v_{j,\;j\in[1,m]}$ is:
\begin{small}
\begin{equation} \label{eq:variabilityWeight}
\frac{\alpha(\nabla(\mathbb{T}_S, \mathbb{T}_T)|_{\mathbb{O}}) + \beta(\nabla(\mathbb{T}_S, \mathbb{T}_T)|_{\mathcal{Y}})}{\alpha + \beta}
\end{equation}
\end{small} 
\end{defn}

\begin{ex}{\label{ex:Weight}}\textbf{(Variability Vector)}\\
Applying \eqref{eq:variabilityWeight} on the variability of semantic learning tasks between $\mathbb{T}_S$ and $\mathbb{T}_T$: $(\sfrac{2}{3}, 0)$ in Example \ref{ex:domainVariability} 
results in ${\bf{v}}(\mathcal{G}, \alpha, \beta) = \sfrac{1}{3}$, which represents moderate variability.
\end{ex}

\subsection{Boosting for Semantic Transfer Learning}

\cjy{Algorithm \ref{algo:StAdaBoost} presents an extension of the transfer learning method TrAdaBoost \cite{dai2007boosting} by integrating semantic embeddings.}
%
\todo{It aims at learning a predictive function $f_{T|_{\mathcal{S}}}(\cdot)$ (line \ref{algo:StAdaBoost:output}) using 
$\langle \mathbb{T}_S, \mathbb{O}_S\rangle$,
$\mathbb{O}_T$ for $\mathbb{T}_T$.}
%
The semantic embeddings of all entailments in $\mathcal{G}_S\cup\mathcal{G}_T$ are computed (lines \ref{algo:StAdaBoost:se1}-\ref{algo:StAdaBoost:se2}). 
They are defined through transferability, consistency, variability effects from source to target domain.
Then, their importance / weight ${\bf w}$ are iteratively adjusted (line \ref{algo:StAdaBoost:foreach}) depending on the evaluation of $f^t$ (lines \ref{algo:StAdaBoost:error1}-\ref{algo:StAdaBoost:error2}) when comparing estimated prediction $f^t({\bf e}_i)$ and real values $\mathcal{Y}_T(g_i)$. 

%
The base model (lines \ref{algo:StAdaBoost:L1}-\ref{algo:StAdaBoost:L2}), which can be derived from any weak learner e.g., Logistic Regression, 
is built on top of all entailments in source,
target tasks.
However, entailments from the source might be wrongly predicted due to tasks  variability 
(Definition \ref{defn:domainVariability} - line \ref{algo:StAdaBoost:se2}) $\mathbb{T}_S$,
$\mathbb{T}_T$.
Thus, we follow the parameterization of $\gamma$ and $\gamma_t$  \cite{dai2007boosting} by decreasing the weights of such entailments 
to reduce their effects (lines \ref{algo:StAdaBoost:w1}-\ref{algo:StAdaBoost:weight}).
%
In the next iteration, the misclassified source 
entailments, which are dissimilar to the target ones w.r.t. semantic embeddings, will affect the learning process less than the current iteration. 
%
Finally, \texttt{StAdaB} returns a binary hypothesis (line \ref{algo:StAdaBoost:output}). Multi-class classification can be easily applied.

%
\begin{algorithm}[h!]
\small

\KwIn{
%
\todo{
(i) Source/target learning domains and tasks $\langle \mathcal{D}_S, \mathbb{T}_S \rangle$, $\langle \mathcal{D}_T, \mathbb{T}_T \rangle$,
%
(ii) a training LSO set of the target learning domain $\mathbb{O}_T^{\prime}$,
(iii) distinct ABox entailments $\mathcal{G}=\{g_1,\ldots,g_m\}$ of $\mathbb{O}_S\cup\mathbb{O}_T^{\prime}$,
}
%
%
(iv) a base learning algorithm ${\bf L}$,
%
(v) 
max. iterations $N$,
(vi) $\alpha, \beta\in[0,1]$.
}
\KwResult{\todo{$f_{T|_{\mathcal{S}}}$: A predictive function by $ \mathcal{D}_S, \mathbb{T}_S$,
$\mathbb{O}_T^{\prime}$, $\mathcal{G}_T^{\mathcal{Y}}$ for $\mathbb{T}_T$.
}}

\Begin{
\emph{\% Initialization of ${\bf w}$eights for transferability, consistency,\nllabel{algo:StAdaBoost:init1}}\\ 
\emph{\% and variability vectors of all $m$ ABox entailments in $\mathcal{G}$.}\\
Initialization of ${\bf w}^{1} \stackrel{.}{=} (w_1^1,\cdots,w_{3\times m}^1);\nllabel{algo:StAdaBoost:init2}$

\emph{\% Computation of semantic embeddings for all $g_i\in\mathcal{G}$.\nllabel{algo:StAdaBoost:se1}}\\
$\mathbf{e}_i \leftarrow (\mathbf{t}(g_i), \mathbf{c}(g_i), \mathbf{v}(\mathcal{G}, \alpha, \beta)),\forall i\in\{1,\cdots,m\};\nllabel{algo:StAdaBoost:se2}$

\ForEach(\emph{\% Weight computation iteration}){$t = 1,2,...,N$\nllabel{algo:StAdaBoost:foreach}}{

${\bf p}^{t} \leftarrow \sfrac{{\bf w}^{t}}{\sum_{i=1}^{3m}w_i^t};$
\emph{\% Probability distribution of ${\bf w}^{t}$.}\\

\emph{\% Predictive function $f^t$ over $\mathbb{O}_S\cup\mathbb{O}_T^{\prime}$.}\nllabel{algo:StAdaBoost:L1}\\
$(f^t: {\bf e}_i \rightarrow \mathcal{Y}_T({\bf e_i})) \leftarrow {\bf L}({\bf e}, {\bf p}^{t}, \mathcal{Y}_T);$\nllabel{algo:StAdaBoost:L2}\\

\emph{\% Error computation of $f^t$ on $\langle \mathbb{T}_T, \mathbb{O}_T^{\prime}\rangle$.\nllabel{algo:StAdaBoost:error1}}\\
$\psi_t \leftarrow {\sum_{i|{{g}_i}\in\mathcal{G}_T}}\frac{w_i^t \cdot |{f^t(\mathbf{e}_i) - \mathcal{Y}_T(g_i)}|}{\sum_{i|{{g}_i}\in\mathcal{G}_T}};$\nllabel{algo:StAdaBoost:error2}\\

\emph{\% Weights for reducing errors on $\mathbb{T}_T$ over iteration.}\nllabel{algo:StAdaBoost:weight1}\\
$\gamma_t \leftarrow \psi_t / (1-\psi_t);\;\; \gamma \leftarrow\sfrac{1}{(1+\sqrt{2\ln(\sfrac{|\mathcal{G}_S|}{N}}))};$ 

\emph{\% Weight update of source and target entailments in $\mathcal{G}$.}\nllabel{algo:StAdaBoost:w1}\\
\emph{\% using $\gamma_t$, $\gamma$, and results from previous iteration: $w_i^t$.}\\
\begin{small}
\begin{equation}\label{eq:g1}
\hspace*{-0.8cm}w_i^{t+1} \leftarrow 
\begin{cases} 
w_i^t \cdot \gamma_t^{-|f^t(\mathbf{e}_i) - \mathcal{Y}_{T}({\bf e}_i)|}, & \mbox{if } g_i\in\mathcal{G}_T \\\nonumber
w_i^t \cdot \gamma^{|f^t(\mathbf{e}_i) - \mathcal{Y}_{T}({\bf e}_i)|}, & \mbox{else}
\end{cases}
\end{equation}
\end{small}\nllabel{algo:StAdaBoost:weight}

}

\Return{Hypothesis ensemble:
\begin{equation*}\label{eq:ensemble}
\todo{f_{T|_{\mathcal{S}}}(\mathbf{e})} = 
\begin{cases}
1, & \mbox{if }\prod_{t=\lceil N/2 \rceil}^N \beta_t^{-f_t(\mathbf{e})} \geq \prod_{t=\lceil N/2 \rceil}^N \beta_t^{-\frac{1}{2}} \\
0, & \mbox{else}
\end{cases}
\end{equation*} \nllabel{algo:StAdaBoost:output}
}

}
\caption{{\label{algo:StAdaBoost}}\small{{\tt{StAdaB}}($\langle \mathcal{D}_S, \mathbb{T}_S\rangle, \langle \mathcal{D}_T, \mathbb{T}_T \rangle, \mathbb{O}_T^{\prime}, \mathcal{G}, {\bf L}, N, \alpha, \beta$)}}
\end{algorithm}

%
A brute force approach would consist in generating 
an exponential number of models with any combination of entailments from 
source,  
target. {\tt{StAdaB}} 
reduced its complexity by only evaluating atomic impact and (approximately) computing the optimal combination. 
As a side effect, 
{\tt{StAdaB}} exposes entailments in the source 
which are driving transfer learning (cf. 
final weight assignment of 
embeddings).


%
%
%
%

%
\section{Experimental Results}\label{sec:evaluation}


%
{\tt{StAdaB}} is evaluated
%
\todo{by two {\bf I}ntra-domain transfer learning cases: 
\textit{(i)} air quality forecasting from {\bf B}eijing to {\bf H}angzhou (IBH),
\textit{(ii)} traffic condition prediction from {\bf L}ondon 
to {\bf D}ublin (ILD), 
one {\bf I}nter-domain case:
\textit{(iii)} from traffic condition prediction in {\bf L}ondon to air quality forecasting in {\bf B}eijing (ILB).
}
Accuracy with cross validation is reported.
%
All tasks are performed with a respective value of $.3$, $.4$, 
$.7$ for variability ${\bf v}(\mathcal{G},\alpha,\beta)$.
%
$\alpha$ and $\beta$ are set to $.5$.

\textbf{
IBH\footnote{Air quality data: https://bit.ly/2BUxKsi. See more about the application and data in \cite{chen2015smog}.}:} 
Air quality knowledge in Beijing (source) is transferred to Hangzhou (target) for forecasting air quality index, ranging 
from Good (value $5$), Moderate ($4$), Unhealthy ($3$), Very Unhealthy ($2$), Hazardous ($1$) to Emergent ($0$).
The observations include air pollutants (e.g., $\text{PM}_\text{2.5}$), meteorology elements (e.g., wind speed) and weather condition from $12$ stations.
The semantics of observations is based on a DL $\mathcal{ALEH(D)}$ ontology, including $48$ concepts, $15$ roles, $598$ axioms.
%
$1,065,600$ RDF triples are generated on a daily basis. $18$ (resp. $5$) months of observations are used as training (resp. testing).
%
\fl{Even though the ontologies are from the same domain, the proportion of similar concepts and roles are respectively $.81$ (i.e., $81\%$ of concepts are similar) and $.74$. For instance, no hazardous air quality concept in Hangzhou.}

\textbf{
ILD:} 
Bus delay knowledge in London (source) is transferred to Dublin (target) for predicting traffic conditions classified as Free (value 4), Low (3), Moderate (2), Heavy (1), Stopped (0).
Source and target domain data 
include bus 
location, delay, congestion status,
weather conditions.
We 
enrich the data
using a DL $\mathcal{EL}^{++}$ domain ontology ($55$ concepts, $19$ roles, $25,456$ axioms).
$178,700,000$ RDF triples are generated on a daily basis. $24$ (resp. $8$) months of observations are used as training (resp. testing).
\fl{The concept and role similarities among the two ontologies are respectively $.73$ and $.77$.}

\textbf{
ILB:} 
\change{Bus delay knowledge in London (source) is transferred to a very different domain: Beijing (target) for forecasting air quality index.
Data and ontologies from IBH and ILD are considered.
Both domains share some common and conflicting knowledge. Inconsistency might then occur.
\fl{For instance, both domains have the concepts of City, weather such as Wind but are conflicting on their importance and impact on the targeted variable i.e., bus delay in London and air quality in Beijing.}
\fl{The concept and role similarities among the two ontologies are respectively $.23$ and $.17$.}
}

%

%
\subsection{Semantic Impact}
Table~\ref{res:Semantics} reports the 
impact of considering semantics (cf. \change{Sem. vs.} Basic) and 
(in)consistency (cf. Consistency / Inconsistency) \change{in semantic embeddings} on Random Forest (RF), Stochastic Gradient Descent (SGD), AdaBoost (AB).  
``\emph{Basic}" models are models with no semantics attached.
``\emph{Plain}" models are modelling and prediction in the target domain i.e., no transfer learning, while ``\emph{TL}" 
refers to transferring entailments from the source.
%
As expected semantics positively boosts accuracy of transfer learning for intra-domain cases (IBH and ILD) with an average improvement of $13.07$\% across models. 
More surprisingly it even over-performs in the inter-domain case (ILB) with an improvement of $20.03$\%.
\todo{
Inconsistency has shown to drive below-baseline accuracy. On the opposite results are much better when considering consistency for intra-domain cases ($63.55\%$), and inter-domain cases ($187.89$\%).
}

\begin{table}[h]
\scriptsize{
\centering
%
\begin{tabular}[bt]{|p{0.35cm}<{\centering}|p{0.15cm}<{\centering}@{ }|@{ }l@{ }|@{ }|c|c||c|c||c|c|}\hline
\multirow{2}{*}{Case} & \multicolumn{2}{|c||}{\multirow{2}{*}{Models}} & \multicolumn{2}{|c||}{RF} & \multicolumn{2}{|c||}{SGD} & \multicolumn{2}{|c|}{AB}\\
& \multicolumn{2}{|c||}{} & \emph{Plain} & \emph{TL} & \emph{Plain} & \emph{TL} & \emph{Plain} & \emph{TL} \\ \hline \hline

\multirow{5}{*}{\begin{sideways}IBH\end{sideways}}& \multicolumn{2}{|c||}{\emph{Basic}} & $.61$ & $.61$ & $.59$ & $.62$ & $.59$ & $.63$\\ \cline{2-9}
& \multirow{3}{*}{\begin{sideways}Sem.\end{sideways}} & Consistency & $.65$ & $.74$ & $.62$ & $.69$ & $.64$ & $.73$\\ \cline{3-9}
&  & Inconsistency & $.56$ & $.64$ & $.52$ & $.60$ & $.49$ & $.63$\\ \cline{3-9}
&  & {\bf Cons. / Incons.} & \multicolumn{2}{|c||}{{\bf \Plus16.07\%}} & \multicolumn{2}{|c||}{{\bf \Plus19.23\%}} & \multicolumn{2}{|c|}{{\bf \Plus30.61\%}}\\ \cline{2-9}
&  \multicolumn{2}{|c||}{{\bf Semantic / Basic}}  & \multicolumn{2}{|c||}{{\bf \Plus13.93\%}} & \multicolumn{2}{|c||}{{\bf \Plus8.18\%}} & \multicolumn{2}{|c|}{{\bf \Plus12.17\%}}\\ \hline \hline
\multirow{5}{*}{\begin{sideways}ILD\end{sideways}}& \multicolumn{2}{|c||}{\emph{Basic}} & $.68$ & $.71$ & $.57$ & $.62$ & $.63$ & $.69$\\ \cline{2-9}
& \multirow{3}{*}{\begin{sideways}Sem.\end{sideways}} & Consistency &  $.75$ & $.78$ & $.65$ & $.71$ & $.75$ & $.82$\\ \cline{3-9}
&  & Inconsistency & $.44$ & $.52$ & $.26$ & $.49$ & $.24$ & $.46$ \\ \cline{3-9}
&  & {\bf Cons. / Incons.} & \multicolumn{2}{|c||}{{\bf \Plus 60.22\%}} & \multicolumn{2}{|c||}{{\bf \Plus102.86\%}} & \multicolumn{2}{|c|}{{\bf \Plus152.35\%}}\\ \cline{2-9}
&  \multicolumn{2}{|c||}{{\bf Semantic / Basic}}  & \multicolumn{2}{|c||}{{\bf \Plus10.07\%}} & \multicolumn{2}{|c||}{{\bf \Plus 14.70\%}} & \multicolumn{2}{|c|}{{\bf \Plus 19.42\%}}\\ \hline \hline
\multirow{5}{*}{\begin{sideways}ILB\end{sideways}}& \multicolumn{2}{|c||}{\emph{Basic}} & $.62$ & $.65$ & $.60$ & $.66$ & $.61$ & $.68$\\ \cline{2-9}
& \multirow{3}{*}{\begin{sideways}Sem.\end{sideways}} & Consistency & $.74$ & $.79$ & $.69$ & $.78$ & $.73$ & $.85$\\ \cline{3-9}
&  & Inconsistency & $.23$ & $.45$ & $.29$ & $.42$ & $.18$ & $.34$\\ \cline{3-9}
&  & {\bf Cons. / Incons.} & \multicolumn{2}{|c||}{{\bf \Plus 153.96\%}} & \multicolumn{2}{|c||}{{\bf 166.25\%}} & \multicolumn{2}{|c|}{{\bf \Plus 243.46\%}}\\ \cline{2-9}
&  \multicolumn{2}{|c||}{{\bf Semantic / Basic}}  & \multicolumn{2}{|c||}{{\bf \Plus 20.44\%}} & \multicolumn{2}{|c||}{{\bf \Plus 17.33\%}} & \multicolumn{2}{|c|}{{\bf \Plus 22.33\%}}\\ \hline
\end{tabular}
\caption{
Forecasting Accuracy / Improvement over State-of-the-art Models (noted as Basic) with Consistency / Inconsistency (Consistency ratio $.8$) based Knowledge Transfer. 
}
\label{res:Semantics}
}
\end{table}

\begin{figure}[h]
\centering
\includegraphics[scale=0.49]{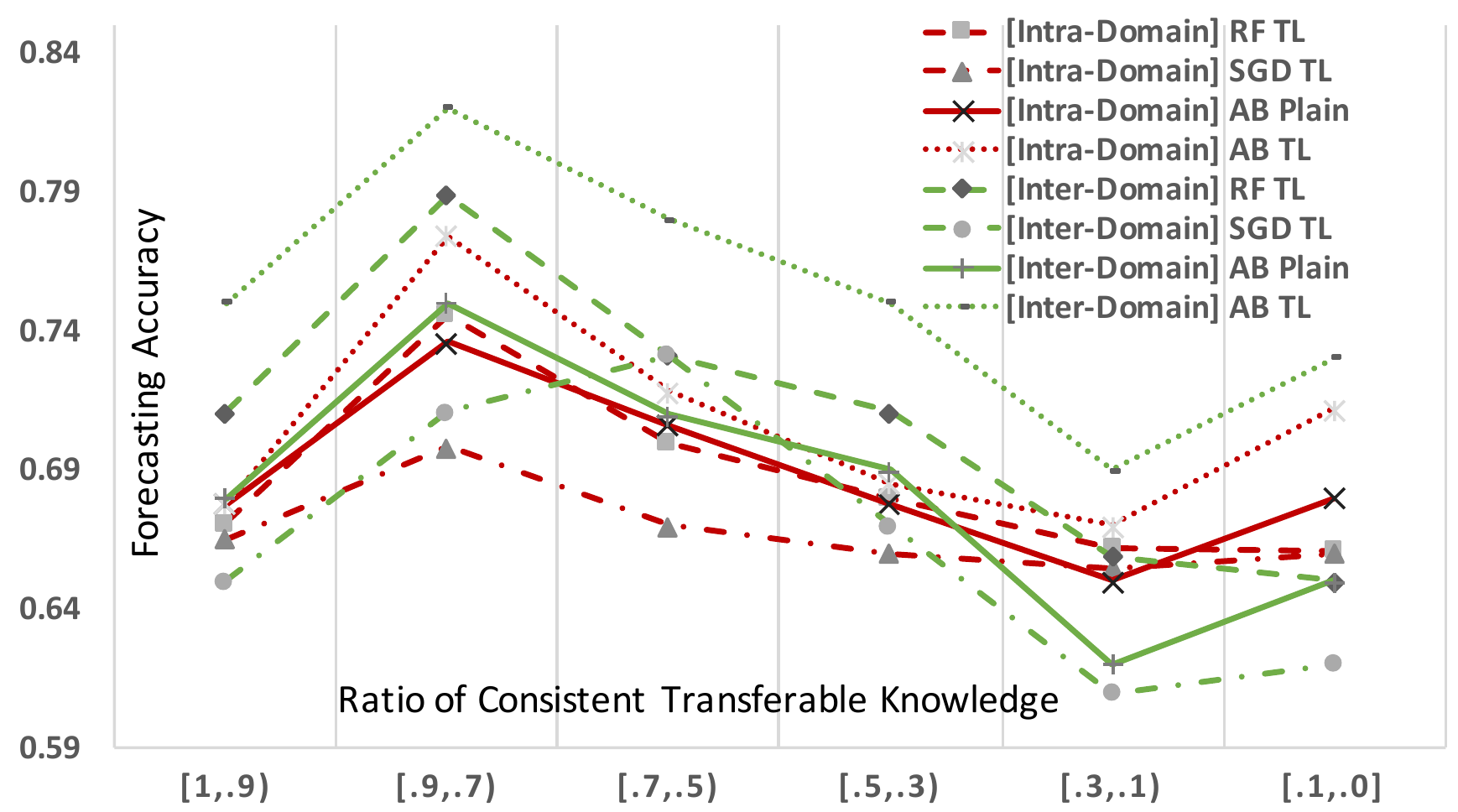}
\caption{Forecasting Accuracy vs. Semantic Consistency.}
\label{res:consistence2}
\end{figure}

%
Figure~\ref{res:consistence2} reports the impact of consistency and inconsistency on transfer learning by analysing how the ratio of consistent transferable knowledge in $[0,1]$ is driving accuracy.
Accuracy is reported for methods in Table \ref{res:Semantics} on intra- (average of IBH and ILD) and inter-domains (ILB).
Max. (resp. min.) accuracy is ensured with ratio in $[.9,.7)$ (resp. $[.3,.1)$).
The more consistent transferable knowledge the more transfer for $[.9,.1)$.
%
Interestingly having only consistent (resp. inconsistent) transferable knowledge 
does not ensure best (resp. worst) accuracy.
\change{This is partially due to under- (resp. over-) populating the target task with conflicting knowledge,}
ending up to limited transferability.
%

%
%

%
\subsection{Comparison with Baselines and Discussion}
\begin{figure}[h]
\centering
\subfigure[Intra-Domain IBH]{
\label{res:baselines_1}
\includegraphics[width=0.225\textwidth]{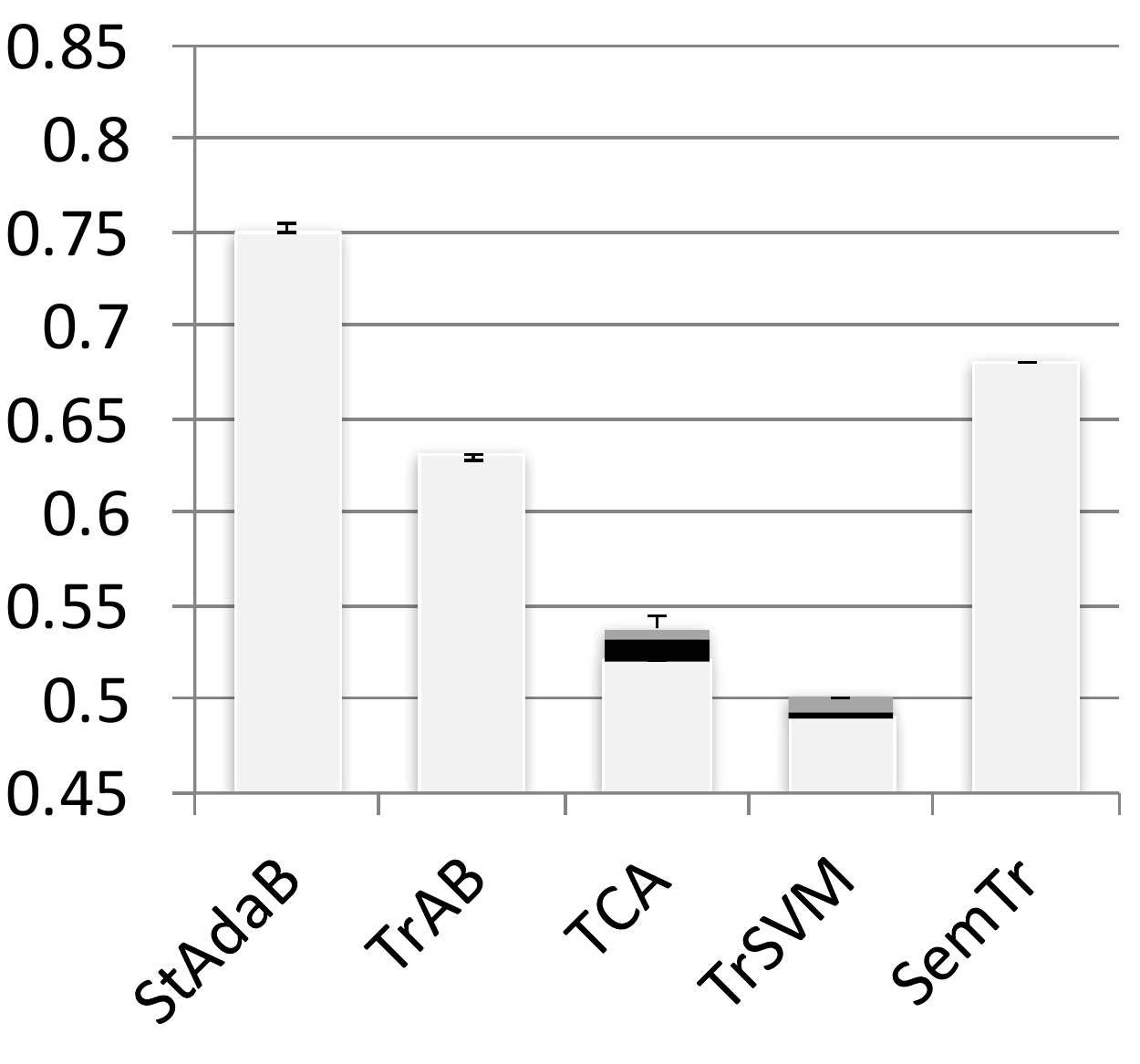}}
\subfigure[Intra-Domain ILD]{
\label{res:baselines_2}
\includegraphics[width=0.225\textwidth]{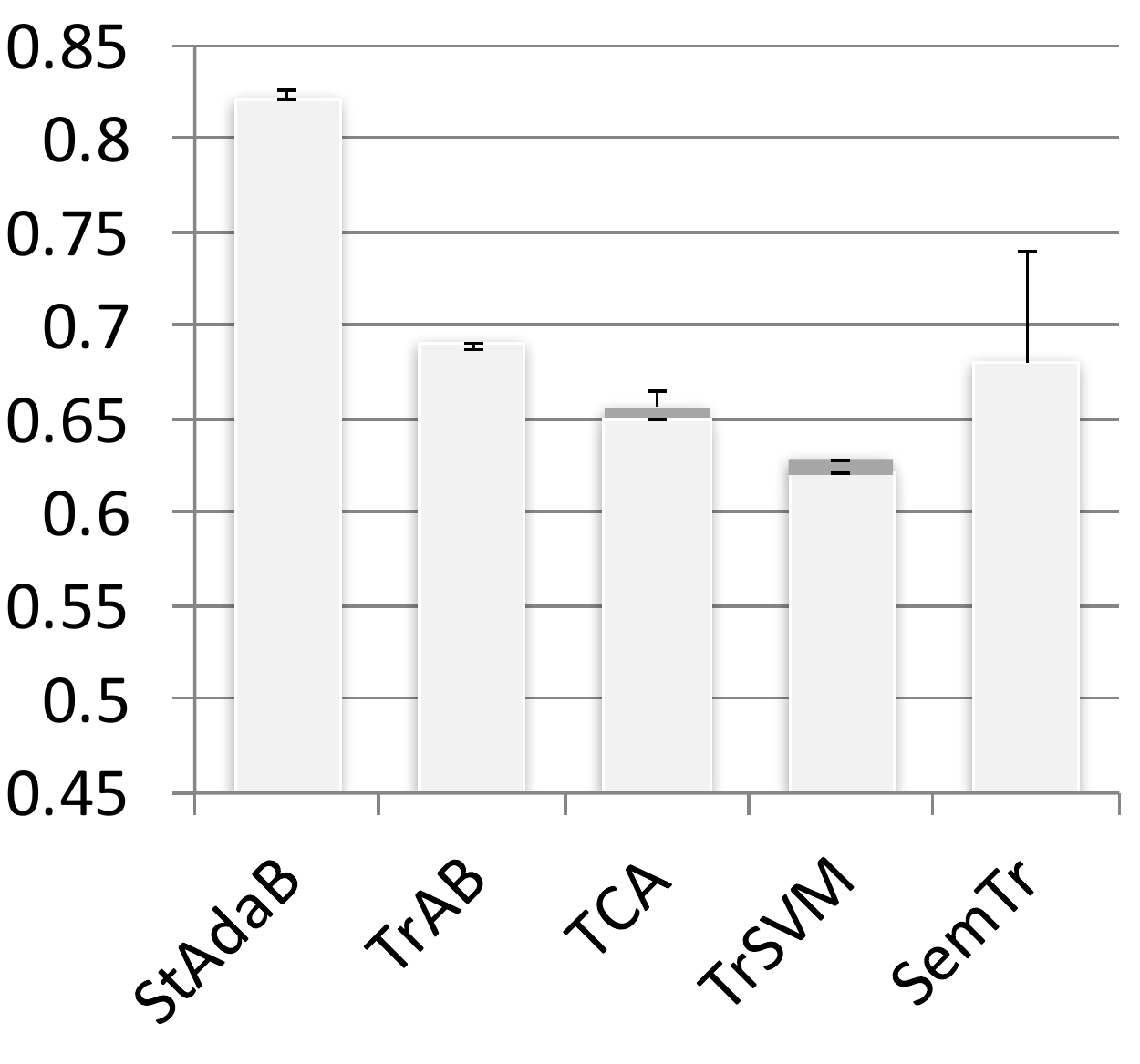}}
\subfigure[Inter-Domains ILB]{
\label{res:baselines_3}
\includegraphics[width=0.225\textwidth]{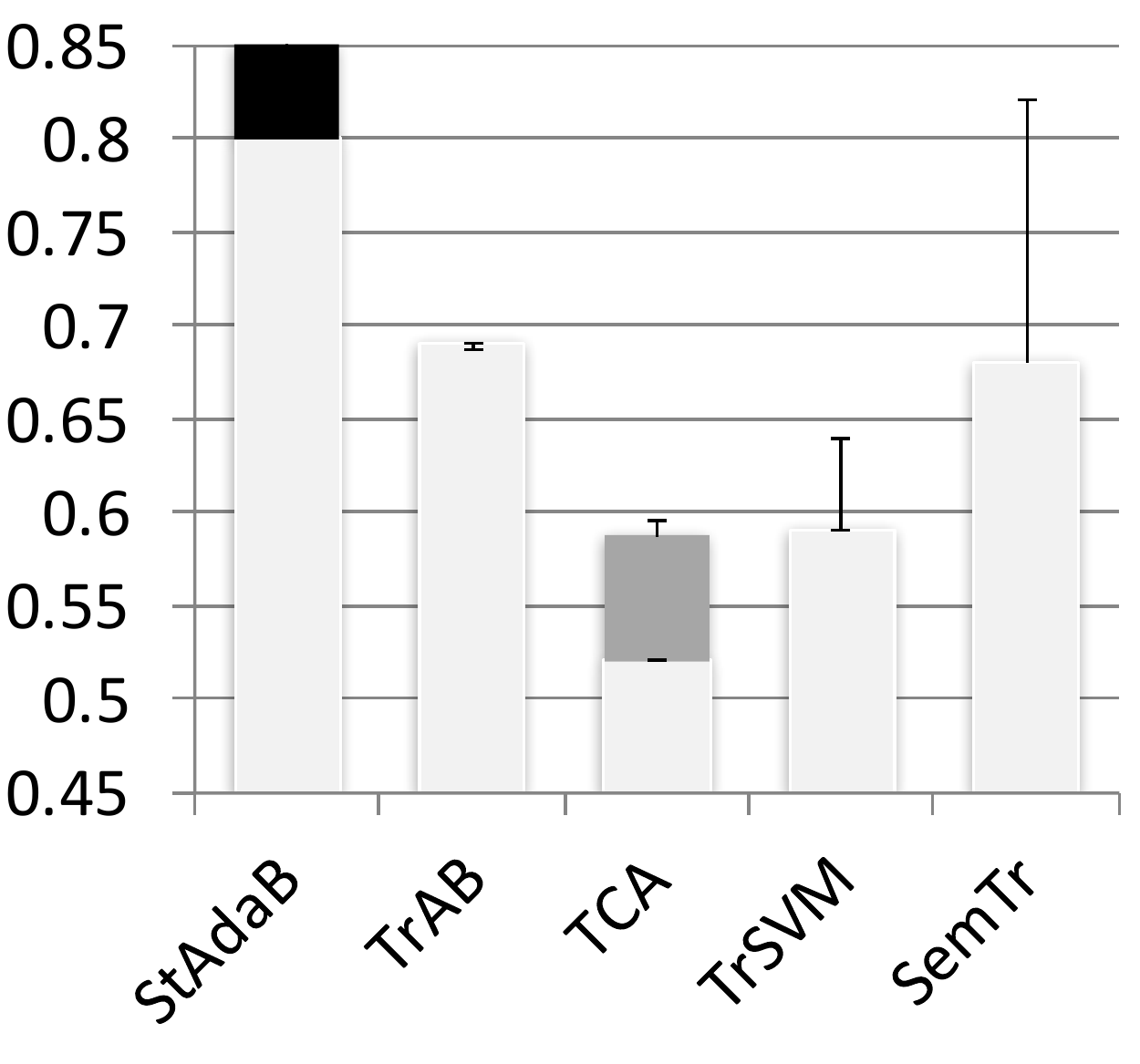}}
\subfigure[Inter-Domains ILB$^*$]{
\label{res:baselines_4}
\includegraphics[width=0.225\textwidth]{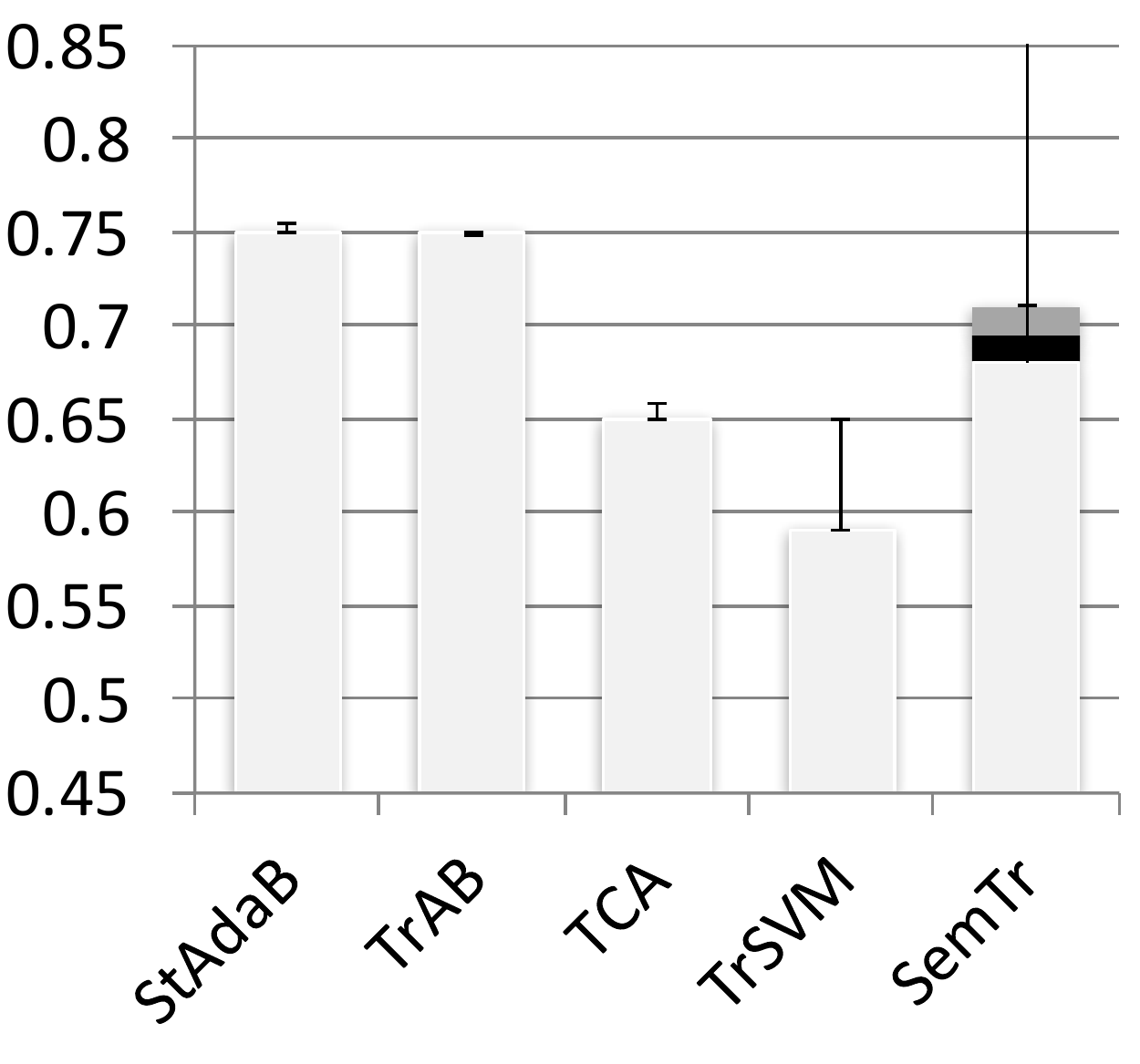}}
\caption{Baseline Comparison of Forecasting Accuracy.
}
\label{res:baselines}
\end{figure}

We compare 
\texttt{StAdaB} (${\bf L}$ $=$ Logistic Regression, $N$ $=$ $800$) with  Transfer AdaBoost TrAB \cite{dai2007boosting},  Transfer Component Analysis (TCA) \cite{pan2011domain},  TrSVM \cite{benavides2017accgensvm} and SemTr \cite{lv2012transfer} (cf. details in Section \ref{sec:RW}).
%
We considered intra-domains: IBH, ILD and inter-domains: ILB and ILB$^*$ (i.e., ILB with same level of semantic expressivity covered by SemTr).
Results report that transfer learning has limitations in the Beijing - Hangzhou context cf. Figure \ref{res:baselines_1}. Although our approach over-performs other techniques (from $10.29\%$ to $50\%$), accuracy does not exceed $74\%$. 
The latter is due to the context, which is limited by the (i) semantic expressivity and (ii) data availability in Hangzhou. 
The results show that TrSVM and TCA reach similar results (average difference of $9.1\%$) in all the cases. 
However our approach and TrAB tend to maximise the accuracy specially in inter-domains ILB and ILB$^*$ in Figures \ref{res:baselines_3} and \ref{res:baselines_4} as both favour heterogenous domains by design.
Interestingly the semantic context of ILB$^*$ in Figure \ref{res:baselines_4} (i) does not favour SemTr much ($\Plus 7.46\%$ vs. ILB), (ii) does not have impact for \texttt{StAdaB} compared to ILB, and more surprisingly (iii) does benefit TrAB ($\Plus 9.15\%$ vs. ILB).
%
This shows that expressivity of semantics is crucial in our approach to benefit from (in-)consistency in transfer.

%

%
%
Adding semantics to domains for transfer learning has clearly shown the positive impact on accuracy, specially in context of inter-domains transfer. This demonstrates the robustness of models supporting semantics when common / conflicting knowledge is shared.
The expressivity of semantics has also shown positive impacts, specially when (in-)consistency can be derived from the domain logics, although some state-of-the-art approaches benefit from taxonomy-like knowledge structure. 
Our approach also demonstrates that the more semantic axioms the more robust is the model and hence the higher the accuracy cf. Figure \ref{res:baselines_1} vs. \ref{res:baselines_2}.
%
%
Data size and axiom numbers are critical as they drive and control the semantics of domain and transfer, which improve accuracy, but not scalability (not reported in the paper). 
It is worst with more expressive DLs due to consistency checks, and with limited impact on accuracy. 
\fl{Enough training data in the source domain is required. Indeed logic reasoning could not help if important data or features are not mapped to the ontology. 
This is crucial for training and validation of semantics in transfer learning. 
Our approach is as robust as other transfer learning approaches, it only differentiate on valuing the transferability at semantic level.}


%
\section{Related Work}{\label{sec:RW}}
We briefly divide the related work into instance transfer, model transfer and semantics transfer.
%
\cjy{Instance transfer selectively reuses source domain samples with weights
%
%
\cite{dai2007boosting}.
%
}
\cite{tan2017distant} select data points from intermediate domains to obtain smooth transfer between largely distant domains.
Model transfer reuses model parameters like features in the target domain.
\cjy{
%
%
For example,
\cite{pan2011domain} introduced a
transfer component analysis for domain adaption;
}
%
%
\cite{benavides2017accgensvm} selectively shares the hypothesis components learnt by Support Vector Machines. 
These methods however usually ignore data semantics.

%
Semantics transfer incorporates external knowledge to boost the above \jp{two} groups,
by using semantic nets~\cite{lv2012transfer}  
or knowledge graph-structure data~ \cite{lee2017transfer}    to derive similarity in data and features, 
%
\jp{
with  no reasoning 
applied. There are  efforts on  Markov Logic Networks (MLN) based transfer learning, by using  first~\cite{MHM2007,MiMo2009} or second order~\cite{DaDo2009,HKD2015} rules as declarative  prediction models. However, these approaches do not address the problem of “when is feasible to transfer”. Our approach uses OWL reasoning to select transferable samples (addressing `when to transfer'),  then enriching the  samples with embedded transferability semantics. 
It can 
support different machine learning models (and not just rules).
}

%
\section{Conclusion}

\change{
We addressed the problem of transfer learning in expressive semantics settings, by
 exploiting semantic variability, transferability and consistency \cjy{to deal with \textit{when to transfer and what to transfer},
for existing instance-based transfer learning methods.} 
It has been shown to be robust to both intra- and inter-domain transfer learning tasks from real-world applications in Dublin, London, Beijing and Hangzhou.
 }
%
As for future work, we will investigate  limits and   explanations of transferability with more expressive semantics, e.g, based on approximate reasoning~\cite{PRZ2016,DPWQ+2019}.

\section*{Acknowledgments}
This work is partially funded by
NSFC91846204. 

\bibliographystyle{named}
\bibliography{reference}

\end{document}